\documentclass[letterpaper, 10 pt, conference]{ieeeconf}  

\IEEEoverridecommandlockouts                              

\usepackage{cite}
\usepackage{url}
\usepackage{hyperref}
\usepackage{amsfonts}
\usepackage{amsmath}
\usepackage{amssymb}
\usepackage{makecell}
\usepackage{multirow}
\usepackage{bm}  
\usepackage{stmaryrd}  
\usepackage{graphicx}  
\usepackage{caption} 
\usepackage{subcaption} 
\usepackage{makecell} 
\usepackage{svg}  
\usepackage{ctable} 

\title{\LARGE \bf
DRL-Based Trajectory Tracking for Motion-Related Modules\\in Autonomous Driving
}

\author{Yinda Xu$^{1}$ and Lidong Yu$^{2}$
\thanks{$^{1}$Yinda Xu is with Zhejiang University,
        No. 38, Zheda Road, Hangzhou, China
        {\tt\small yinda\_xu@zju.edu.cn}}%
\thanks{$^{2}$Lidong Yu is with Beijing Institute of Technology,
        No. 5, South Street, Zhongguancun, Haidian District, Beijing, China
        {\tt\small yvlidong@gmail.com}}%
}

\begin{document}

\maketitle
\thispagestyle{empty}
\pagestyle{empty}

\begin{abstract}





Autonomous driving systems are always built on motion-related modules such as the planner and the controller. An accurate and robust trajectory tracking method is indispensable for these motion-related modules as a primitive routine.
Current methods often make strong assumptions about the model such as the context and the dynamics, which are not robust enough to deal with the changing scenarios in a real-world system.
In this paper, we propose a Deep Reinforcement Learning (DRL)-based trajectory tracking method for the motion-related modules in autonomous driving systems. The representation learning ability of DL and the exploration nature of RL bring strong robustness and improve accuracy. Meanwhile, it enhances versatility by running the trajectory tracking in a model-free and data-driven manner. Through extensive experiments, we demonstrate both the efficiency and effectiveness of our method compared to current methods. Code~\footnote[3]{Python RL Training and C++ SDK: \url{https://github.com/MARMOTatZJU/drl-based-trajectory-tracking}} and documentation~\footnote[4]{Documentation: \url{https://drl-based-trajectory-tracking.readthedocs.io}} are released to facilitate both further research and industrial deployment.


\end{abstract}

\section{INTRODUCTION}



A reliable autonomous driving system needs a set of effective and efficient motion-related modules to deal with complex scenarios. Trajectory tracking~\cite{hoffmann2007autonomous} is one of the key components and is used in motion-related modules such as the planner~\cite{chen2019autonomous} and the controller~\cite{ivanovic2020mats, liniger2015optimization}. Currently trajectory trackers are either based on the \textit{heuristics model}~\cite{coulter1992implementation,hoffmann2007autonomous}, or  \textit{the trajectory optimization}~\cite{chen2019autonomous, gu2017improved}. The heuristics model leverages the hand-crafted policy to give an efficient performance and cover a certain part of scenarios. But it has difficulties handling speed distribution with a huge diversity and reference lines with complex shapes. In contrast, the trajectory optimization achieves high tracking precision with a proper initial trajectory, but an unsuitable initial trajectory will lead to a failure due to the non-linear nature of the dynamics~\cite{boyd2004convex}.





We argue that the main problem of both approaches comes from the dependency on an accurate and stationary model. Such dependency is hard to be satisfied in real-world applications on various dimensions, such as the dynamics model and task context. For example, the wheelbase and the position of the Center of Mass (CoM) may slightly differ from their ground truths due to measurement errors. Besides, the heuristics designed for a low initial speed cannot handle a high initial speed. In addition, heuristics and optimization-related elements (e.g. jacobian and cost weights) are sensitive to the variation of the model, which reduces robustness, and further accuracy and versatility~\cite{rajamani2011vehicle, boyd2004convex}.

\def\IllustrationSubfigureWidth{0.22\textwidth}

\begin{figure}
\vspace{+0.2cm}  
\centering
 \begin{subfigure}[b]{\IllustrationSubfigureWidth}
     \centering
     \includegraphics[scale=1.0,width=\textwidth]{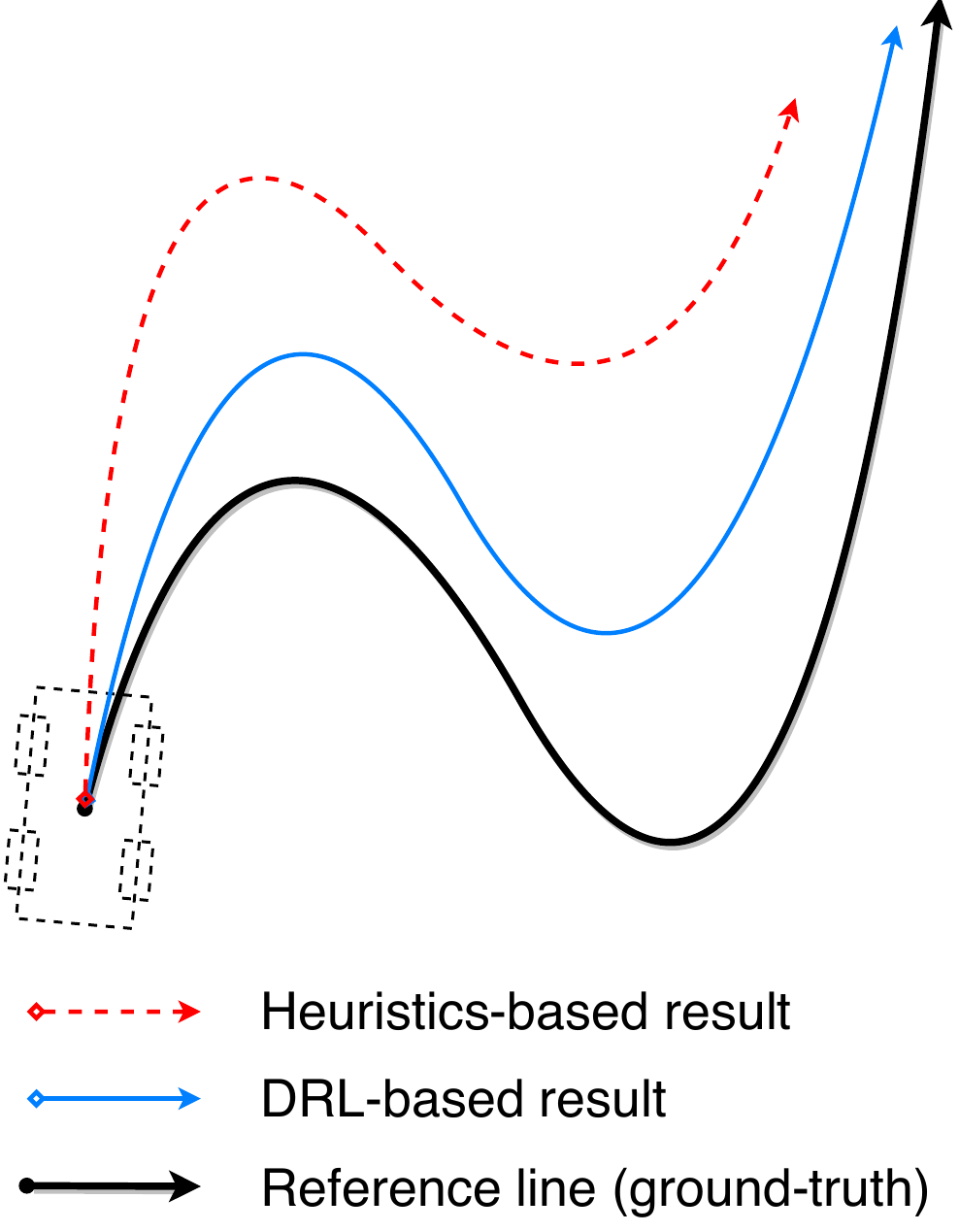}
     \caption{Results before post-optimization}
 \end{subfigure}
 \hfill
 \begin{subfigure}[b]{\IllustrationSubfigureWidth}
     \centering
     \includegraphics[scale=1.0,width=\textwidth]{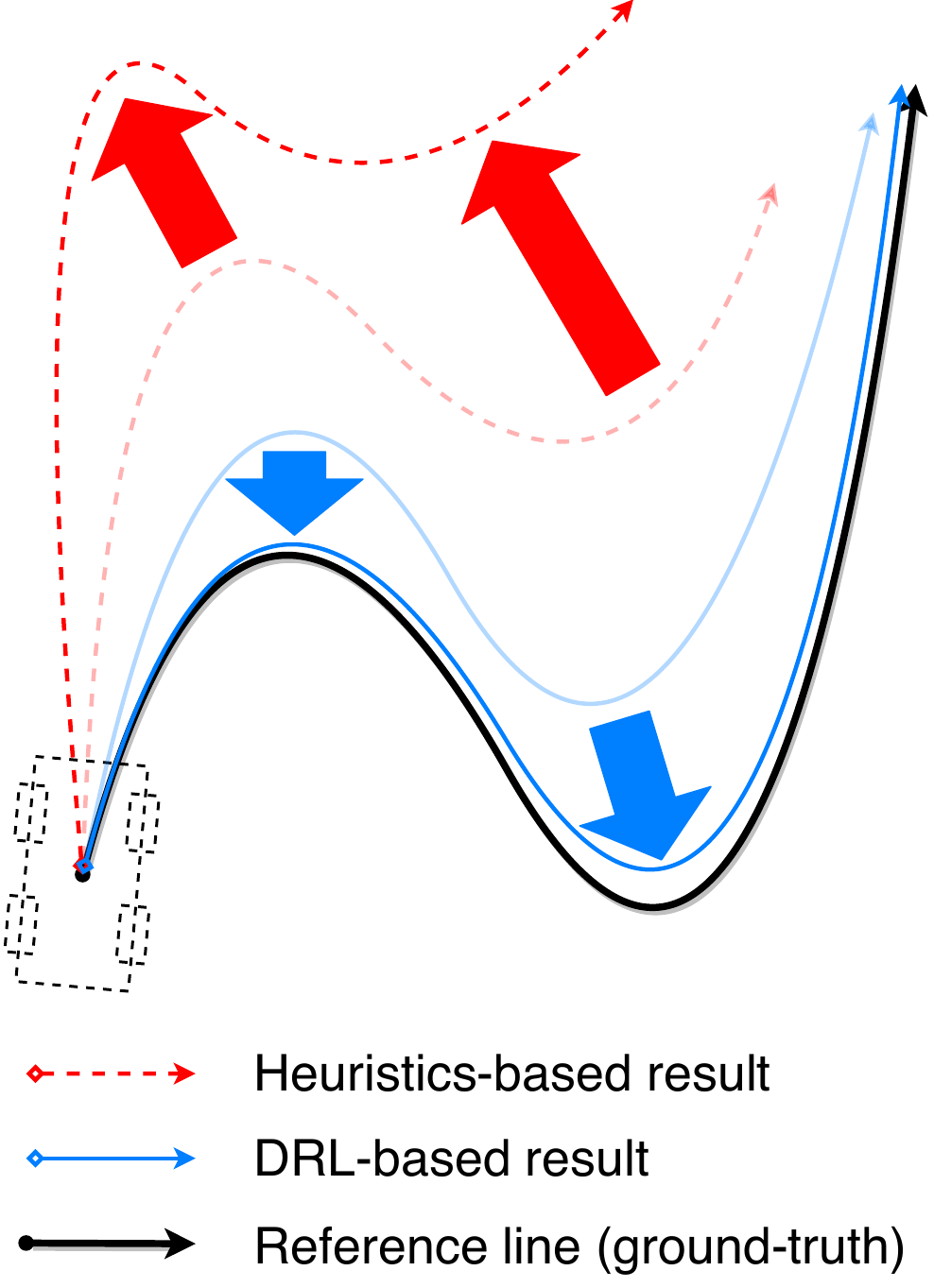}
     \caption{Results after post-optimization}
 \end{subfigure}
\caption{Idea illustration.}
\label{figure:illustration}
\end{figure}

In this paper, we propose to perform the trajectory tracking task via the Deep Reinforcement Learning~\cite{sutton2018reinforcement} (DRL). Figure~\ref{figure:illustration} illustrates the performance gap between the heuristics-based method and the proposed DRL-based method. Our method consists of a trajectory tracking stage and a trajectory optimization stage.
Compared to the heuristics-based method, the DRL method is more robust in the first stage and provides a proper initial trajectory for the post-optimization stage  to coverage in fewer steps.
Our method provides stronger robustness with the representation learning ability of Neural Networks (NN) and Deep Learning (DL)~\cite{goodfellow2016deep}, and the exploration nature of Reinforcement Learning (RL)~\cite{sutton2018reinforcement}.
We tackle the problem of dependency on the model by making the least assumptions and thus introduce accuracy and versatility.

Extensive experiments are conducted to validate the effectiveness of our method. For trajectory tracking under $v_\text{init}$ = 25 [m/s], the drl-L tracker achieves 77\% relative reduction in error, which has already outperformed the two-stage result of the heuristics-based method with additional two post-optimization steps.

\section{BACKGROUND AND FORMULATION}

\subsection{MDP Formulation}



In this work, we formulate the trajectory tracking as a discrete-time Markov Decision Process (MDP)~\cite{sutton2018reinforcement}. 
Within an MDP, an ego-agent traverses a state space, executes actions based on a certain policy,  and accumulates rewards at each step.
An MDP can be represented by $\langle \mathcal{S}, \mathcal{A}, \mathbf{T}, \mathbf{R} \rangle$ whose elements indicate the state space, the action space, the transition function (the dynamics model), and the reward function, respectively. Besides, we treat the discount factor $\gamma$ as a hyper-parameters of the environment. 

An ego-agent needs to observe the state under the world coordinate system and build an ego-centric descriptor under the body coordinate system viewed from this state.
We formulate this process with an observation space $\mathcal{O}$ along with an observation function $\mathbf{O}: \mathcal{S} \to \mathcal{O}$~\footnote{Here we consider a full observation rather than a partial observation. The latter is usually studied under the formulation of Partially Observed MDP (POMDP), which involves the maintenance and updates of the belief of the state and is beyond the scope of this work.}.

Under the framework of MDP, a policy $\Pi: \mathcal{O} \to \mathcal{A}$ maps the current observation $o \in \mathcal{O}$ to an action $a \in \mathcal{A}$, i.e. $a = \Pi(o)$. The resulting action is then executed at the current timestep $t$ and the state transition is described by $s_{t+1} = \mathbf{T}(s_t, a_t)\in \mathcal{S}$. By rolling out the transition function and the policy alternatively, a state sequence or trajectory with length $T$ is yielded at the end of the tracking as
\begin{equation}
\begin{array}{ll}
    \tau = & \{s_{t} \mid t=0, 1, ... T-1\} \\
    \textnormal{s.t.} & o_{t} = \mathbf{O}(s_{t}) \\ & a_{t} = \Pi(o_{t}) \\ & s_{t+1} = \mathbf{T}(s_{t}, a_{t})
\end{array}
.
\end{equation}

\subsection{Trajectory Tracking}
\label{subsec:trajectory-tracking}

The primary objective of the trajectory tracking task is to find a policy that drives the agent to track the reference trajectory with minimum tracking errors. Formally, we have
\begin{equation}
\label{eq:tracking-objective}
\begin{array}{c}
J = \sum_{t=0}^{T-1} \Vert z_{t} - z_{t}^{\star} \Vert = \sum_{t=0}^{T-1} \Vert \mathbf{Z}\left(s_{t} \right) - z_{t}^{\star} \Vert \\
s_{t} \in \mathcal{S}, z_{t}^{\star} \in \mathcal{Z}
\end{array}
,
\end{equation}
where $\Vert \cdot \Vert$ denotes the norm and can be configured differently across various scenarios, $z_t^{\star} \in \mathcal{Z}$ is the waypoint of the tracking target $\mathbf{z} = \{ z_{t} \mid t=0, 1, ... T-1 \}$, i.e. the reference line. The other objectives include energy efficiency, collision-free property, the comfort of passengers, etc. The usage of these objectives for our reward is discussed in Section~\ref{subsec:reward_design} later.

In real-world applications, the waypoints of reference trajectory seldom contain full state information due to task requirements, partial observability, unstable observation quality, etc. For example, when using a bicycle model~\cite{rajamani2011vehicle}, a reference waypoint often consists of the position coordinates without the heading and the velocity. Thus we define the reference state space $\mathcal{Z}$ with a reference state model $\mathbf{Z}: \mathcal{S} \to \mathcal{Z}$ mapping from the state space to the reference waypoint space. For the rest part of this article, we assume reference states are points on $\mathbb{R}^2$ surface, i.e. $\mathcal{Z} = \mathbb{R}^2$, which fits the need of autonomous driving where reference lines are often given in form of a list of 2-D waypoints.



\subsection{Deep Reinforcement Learning}

The Deep Reinforcement Learning (DRL) community has been going through rapid development in recent years. By leveraging the representation ability of the Neural Network and highly efficient training infrastructure (e.g. PyTorch~\cite{paszke2019pytorch}), the State-of-the-Arts RL algorithms scale to difficult problems through various techniques such as learning to search~\cite{Silver2017MasteringCA}, reconstructing the value space~\cite{mnih2013playing}, directly optimizing the policy in an online manner ~\cite{schulman2017proximal}, and combining the above practical measures~\cite{fujimoto2018addressing,haarnoja2018soft}, etc. The Supervised Learning~\cite{krizhevsky2017imagenet} and the Imitation Learning~\cite{bansal2018chauffeurnet} perform a Maximum Likelihood Estimation (MLE) over the training data and are often limited by the data collecting pipeline. In contrast, RL methods aim at maximizing the reward via a comprehensive exploration of the environment and thus suffer little from out-of-distribution (OOD) data samples. In the following of this work, we demonstrate that our DRL-based method is capable to deliver an accurate and robust trajectory tracker and offer a pipeline with low complexity at the same time.


\section{METHODS}

\label{sec:methods}

In this section, we describe our DRL-based method at the component level and the implementation detail of our DRL-based trajectory tracker.


\subsection{Selection of Dynamics Models}

\begin{table}  
\vspace{+0.2cm}  
\caption{Definitions of dynamics models}
\label{tab:dynamics-model-definition}
\begin{center}
\resizebox{0.48\textwidth}{!}{
\begin{tabular}{|c|c|c|c|}
\hline
Name & Bicycle model & Unicycle model \\
\hline
\hline
\makecell{Application \\ category(ies)} & 
Vehicle / Bicycle & 
Pedestrian \\

\hline

\makecell{State \\ space \\ $\mathcal{S}$} &
$
\begin{array}{cc}
    x & \text{X-coordinate} \\
    y & \text{Y-coordinate} \\
    \theta & \text{heading angle} \\
    v & \text{scalar velocity}\\
\end{array}
$ & 
$
\begin{array}{cc}
    x & \text{X-coordinate} \\
    y & \text{Y-coordinate} \\
    \theta & \text{heading angle} \\
    v & \text{scalar velocity}\\
\end{array}
$ \\

\hline

\makecell{Action \\ space \\ $\mathcal{A}$} &
$
\begin{array}{cc}
    \delta & \text{steering angle} \\
    a & \text{scalar acceleration} \\
\end{array}
$ & 
$
\begin{array}{cc}
    \omega & \text{heading change rate} \\
    a & \text{scalar acceleration} \\
\end{array}
$ \\

\hline

\makecell{Hyper- \\ parameters \\ $\theta_\text{d}$} &
$
\begin{array}{cc}
    l & \text{vehicle length} \\
    l_{\text{fo}} & \text{front overhang} \\
    l_{\text{w}} & \text{wheelbase} \\
    l_{\text{ro}} & \text{rear overhang} \\
    w & \text{vehicle width} \\ 
\end{array}
$ & 
No hyper-parameter \\

\hline

\makecell{Transition \\ function \\ $\mathbf{T}$} &
$
\begin{array}{rl}
    \dot{x} &= v \cdot \cos\left( \theta + \beta \right) \\
    \dot{y} &= v \cdot \sin\left( \theta + \beta \right) \\
    \dot{\theta} &= \frac{v \cdot \sin\left( \delta \right)}{l_\text{w}} \\ 
    \dot{v} &= a \\
    \beta &= \arctan\left( \frac{l_\text{r}}{l_\text{w}} \tan\left( \delta \right) \right) \\ 
    l_\text{r} &= \frac{l}{2} - l_\text{ro}\\
\end{array}
$ & 
$
\begin{array}{rl}
    \dot{x} &= v \cdot \cos\left( \theta \right) \\
    \dot{y} &= v \cdot \sin\left( \theta \right) \\
    \dot{\theta} &= \omega \\
    \dot{v} &= a \\
\end{array}
$
\\

\hline

\makecell{State \\ space \\ range} &
$
\begin{array}{cc}
    x & \in(-\infty, +\infty) \text{[m]} \\
    y & \in(-\infty, +\infty) \text{[m]} \\
    \theta & \in[-\pi, +\pi) \text{[rad]} \\
    v & \in[0, 40] \text{[m/s]}\\
\end{array}
$ & 
$
\begin{array}{cc}
    x & \in(-\infty, +\infty) \text{[m]} \\
    y & \in(-\infty, +\infty) \text{[m]} \\
    \theta & \in[-\pi, +\pi) \text{[rad]} \\
    v & \in[0, 4] \text{[m/s]}\\
\end{array}
$ \\

\hline

\makecell{Action \\ space \\ range} &
$
\begin{array}{cc}
    \delta & \in[-0.52,+0.52] \text{[rad]} \\
    a & \in[-4.5,+4.5] \text{[m/s\textsuperscript{2}]} \\
\end{array}
$ & 
$
\begin{array}{cc}
    \omega & \in[-1.57,+1.57] \text{[rad/s]} \\
    a & \in[-3.0,+3.0] \text{[m/s\textsuperscript{2}]} \\
\end{array}
$ \\

\hline
\end{tabular}
}
\end{center}
\end{table}

Within the MDP formulation, a dynamics model defines the transition path within the state space and is conditioned on the state and action. Under the context of autonomous driving, we consider the bicycle model~\cite{rajamani2011vehicle} and the unicycle model~\cite{batkovic2018computationally}. Table~\ref{tab:dynamics-model-definition} gives a full description of the dynamics model involved in this work.




\subsection{Observation Encoding}
\label{subsec:observation_encoding}


We build a complete observation encoding with NN prediction to ensure effective trajectory tracking. 
Our observation $o = \mathbf{O}(s) = \{o^{\tau}, o^{s}, o^{d} \}$ consists of three parts: local observation of the reference trajectory $o^{\tau}$, pose-invariant state information $o^{s}$, and dynamics model information $o^{d}$.


To form a concise, effective, and pose-invariant description of the tracking target, we slice a segment of reference trajectory $\{ z_{i}^{\star} \}$ of length $T_{o}$ ahead from the current timestep $i$, and transform its coordinates to the body (local) coordinate frame of the agent with a transform matrix $M_{\textbf{world} \mapsto \textbf{body}}$. The encoding of a reference line can be described as
\begin{equation}
\begin{array}{l}
    \mathbf{z}_{i}^{\star} = \{ z_{t} \mid t = i, i+1, ..., i+T_{o}-1 \} \\

    o^{\tau} = \{M_{\textbf{world} \mapsto \textbf{body}} \cdot [z; 1] \mid z \in \mathbf{z}_{i}^{\star} \} \\ 
    M_{\textbf{world} \mapsto \textbf{body}} = 
    \begin{bmatrix}
    \cos{\theta_{i}} & 0 & x_{i}
	\\
    0 & \sin{\theta_{i}} & y_{i}
	\\
    \end{bmatrix}
\end{array}
,
\end{equation}
where $\left(x_i, y_i, \theta_i\right)$ is the agent's pose in the world frame, and $[z; 1]$ is the homogeneous coordinate.


The state information describes the position-invariant motion information of the agent at the current timestep. We use the velocity of the Center of Mass (CoM) of the agent $v_{i}$ as the state information at the timestep $i$, i.e.
\begin{equation}
    o^s = \{ v_{i} \}
\end{equation}

To provide priors related to the state transition. We concatenate all the hyper-parameters $\theta^\text{d}$ of the involved dynamics model to form the dynamics model observation. For example, the observation of a bicycle model is
\begin{equation}
    o^\text{d} = \theta^\text{d} = \{l, l_\text{fo}, l_\text{w}, l_\text{ro}, w\}
.
\end{equation}


\subsection{Domain Randomization during Training Phase}
\label{subsec:domain_randomization}

We mainly employ two domain randomization techniques~\cite{tobin2017domain} to encourage a comprehensive exploration of the environment and ensure the robustness of our policy network. Figure~\ref{figure:domain-randomization} illustrates these processes.

The first domain randomization technique is the random generation of reference lines applied during the training phase. Instead of using data from the real world, we consider a more generic task where we generate the tracking trajectories using random walks. 
We sample an initial state from the state space and iteratively conduct a random walk by sampling actions from the action space before executing it with the dynamics model. After yielding a reference line of the desired length, we set this as the reference line of our training environment. 

Figure~\ref{figure:domain-randomization} illustrates this process, where the random walk process can be formalized as 
\begin{equation}
\begin{array}{l}
    s_{i+1} = \mathbf{T}(s_{i}, a_{i}), a_{i} \sim \mathcal{A}, s_{0} \sim \mathcal{S} \\
    \mathbf{z}^{\star} = \{\mathbf{Z}(s_{i}) \mid i = 0, 1, ... T-1\} \\
\end{array}
.
\end{equation}

Here, we deem the real-world data as a subset of reference lines generated from random walks, which can simply be tracked by a policy converged in our training environment. 
The generated training reference lines raise the coverage difficulty with rapidly changing actions, which makes the learning of the policy network gain robustness.




Another domain randomization technique employed in this work is the random sampling of the hyper-parameters of the dynamics model. 
At the reset stage of each episode, we randomly sample the hyper-parameters of the dynamics model from a predefined distribution $\theta_d \sim \mathcal{D}_{\theta_d}$ and set it for the reference line generation and the experience collection during the current episode. 


As shown in Figure~\ref{figure:domain-randomization}, both the policy network and the value network use the sampled information of the dynamics model during the whole episode. This forces the learning to utilize the hyper-parameters of the dynamics model and thus is able to generalize to different hyper-parameter categories.

\begin{table}  
\vspace{+0.2cm}  
\caption{All parameter combinations used for domain randomization}
\label{tab:dynamics-model-parameters}
\begin{center}
\resizebox{0.4\textwidth}{!}{
\begin{tabular}{|c||c||c||c||c|}
\hline
Vehicle type & \makecell{Front\\overhang\\$l_\text{fo}$ [m]} & \makecell{Rear\\overhang\\$l_\text{ro}$ [m]} & \makecell{Wheelbase\\$l_\text{w}$ [m]} & \makecell{Width\\$w$ [m]} \\
\hline
\makecell{Short vehicle\\(e.g. sedan)} & 0.9 & 0.9 & 2.7 & 1.8 \\
\hline
\makecell{Middle vehicle\\(e.g. light-duty truck)} & 1.095 & 1.54 & 3.360 & 2.648 \\
\hline
\makecell{Long vehicle\\(e.g. bus)} & 2.3 & 2.0 & 6.1 & 2.5 \\
\hline
\end{tabular}
}
\end{center}
\end{table}

To avoid the sophisticated design of $\mathcal{D}_{\theta_d}$, we use a discrete uniform distribution that assigns an equal probability to each hyper-parameter set. Table~\ref{tab:dynamics-model-parameters} shows all hyper-parameter sets for the bicycle model. We do not apply this technique to the unicycle model as it does not have any hyperparameter.

\def\DRSubfigureWidth{0.2\textwidth}

\begin{figure}
\centering
\vspace{+0.2cm}  
 \begin{subfigure}[b]{\DRSubfigureWidth}
     \centering
     \includegraphics[width=\textwidth]{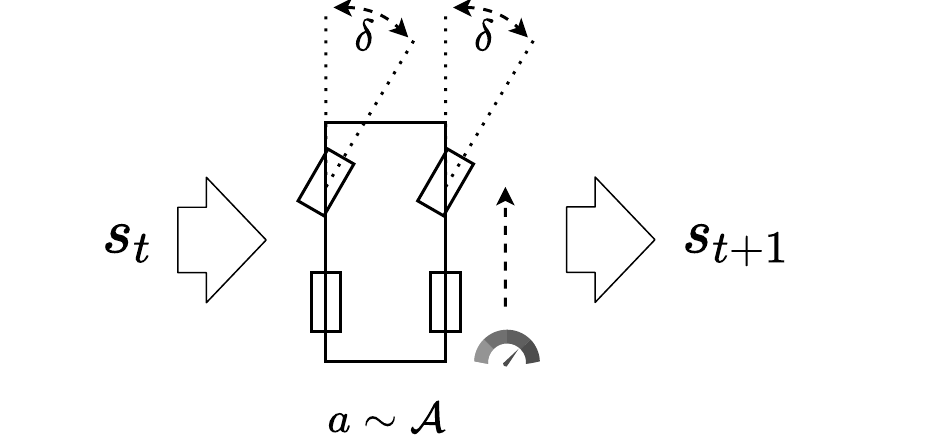}
     \caption{Action sampler}
 \end{subfigure}
 \hfill
 \begin{subfigure}[b]{\DRSubfigureWidth}
     \centering
     \includegraphics[width=\textwidth]{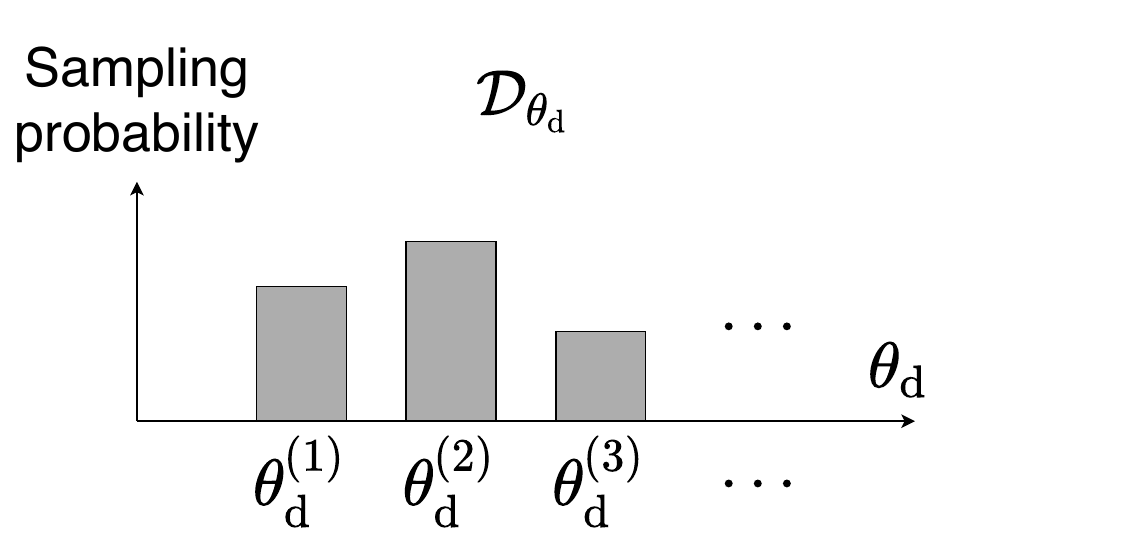}
     \caption{Hyper-parameter distribution}
 \end{subfigure}
 \hfill
 \begin{subfigure}[b]{0.4\textwidth}
     \centering
     \includegraphics[width=\textwidth, trim=1.0cm 0.0cm 0.0cm 0.0cm, clip]{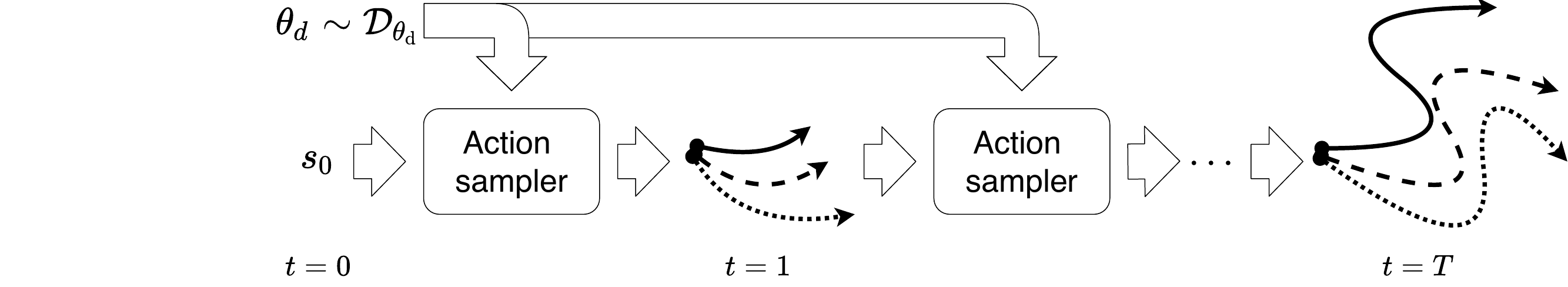}
     \caption{Random walk}
 \end{subfigure}
\caption{Domain randomization techniques. The bicycle model is used for illustration.}
\label{figure:domain-randomization}
\end{figure}


\subsection{Reward Design}
\label{subsec:reward_design}

The reward in DRL serves as a feedback signal from the environment. It assigns a high reward value to an action leading to desired future state given the current state and returns a low reward value for an action in the opposite direction.

The design of the reward function plays a crucial role in the training of the yielded policy. We 
consider two of the most prioritized factors in the trajectory tracking task, i.e. accuracy and smoothness. The joint reward can be indicated as
\begin{equation}
    \mathbf{R}(s) = w^\text{t} \cdot \mathbf{R}^\text{t} + w^\text{a} \cdot \mathbf{R}^\text{a}
,
\end{equation}
where $\mathbf{R}_\text{t}$ and $\mathbf{R}_\text{a}$ denotes respectively the tracking accuracy reward and action reward, weighted by $w_\text{t}$ and $w_\text{a}$.



For the design of $\mathbf{R}_\text{t}$, we consider the tracking objective formalized in Equation~\ref{eq:tracking-objective}. We compute the negative Euclidean distance between the reference waypoint $z_{\star}$ and the agent's position $z$ to form our primary reward as
\begin{equation}
    \mathbf{R}^\text{t} = - \Vert z - z^{\star} \Vert_2^2
\end{equation}
As for the $\mathbf{R}_\text{a}$, to ensure a smooth trajectory, we penalize large action outputs and use a negative norm of action as our action reward by 
\begin{equation}
    \mathbf{R}^\text{a} = - \sum_{j=1}^{\vert\mathcal{A}\vert} \bar{a}_j^2
,
\end{equation}
where $j$ denote the index of action vector and $\vert\mathcal{A}\vert$ indicates the dimensionality of the action space. We use the normalized action value $\bar{a}_j$ with a normalization function chosen by the type of action space. In the autonomous driving scenario, we have a bounded action space and thus the min-max normalization is sufficient, i.e. $\bar{a}_j = \frac{ a_j - a_j^{\text{min}} }{ a_j^{\text{max}} - a_j^{\text{min}} } $.



\subsection{Policy Learning}

We train a policy network that serves as our trajectory tracker by combining an off-the-shelf DRL algorithm with the domain randomization techniques (see Section~\ref{subsec:domain_randomization}).

\subsubsection{Environment and Training}

\label{subsec:environment-and-training}

We build our learning environment based on OpenAI Gym~\cite{brockman2016openai} by implementing the elements described in Section~\ref{sec:methods}. Among the State-of-the-Arts DRL algorithms, we choose TD3~\cite{fujimoto2018addressing} and its implementation in Stable Baselines 3~\cite{raffin2019stable}, which is a PyTorch~\cite{paszke2019pytorch}-based RL library. We use the Multi-Layer Perceptron (MLP)~\cite{haykin1998neural} to build the policy network (actor) and the value network (critic). All layers are activated with ReLU except for the last layer which is instead activated with Tanh to form a bounded output space~\cite{goodfellow2016deep}.
We build two policy networks of different scales, denoted as "drl" and "drl-L" ("L" stands for "large") respectively. Table~\ref{tab:env-rl-hyperparams} lists the details of the environment and the training.

\begin{table}  
\vspace{+0.2cm}  
\caption{
Hyper-parameters of TD3 and the environment
}
\label{tab:env-rl-hyperparams}
\begin{center}
\resizebox{0.49\textwidth}{!}{
\begin{tabular}{|c|c||c|c|}
\hline

\multicolumn{4}{|c|}{TD3 hyper-parameters} \\
\hline

\makecell{Policy network \\ architecture (drl)}  & [128, 32] &
\makecell{Policy network \\ architecture (drl-L)} & \makecell{[256, 256, 128, \\ 128, 64, 64]} \\
\hline

Learning rate & $4.0 \times 10^{-5}$  &
\makecell{Value network \\ architecture} & [1024, 512, 256, 128] \\
\hline

Batch size & 4096 & Training frequency & every 4 steps \\ 
\hline

\makecell{Soft update \\ coefficient}  & 0.002 &
Discount factor & 0.99 \\ 
\hline
Total timesteps & $3 \times 10^6$ & Number of simulations $N_\text{mc}$ & 500 \\
\hline

\hline
\multicolumn{4}{|c|}{Environment hyper-parameters} \\
\hline
Trajectory length  & 5.5 [s] &
Discrete step length & 0.1 [s] \\ 
\hline

\end{tabular}
}
\end{center}
\end{table}

\subsubsection{Observation Encoding}

For the bicycle model, We choose an observation horizon $T_o$ of $13$ steps to strike a balance between the observation-action correlation and the information completeness. 
Each waypoint consists of two coordinates (X and Y), and a 26-D observation vector is formed to encode the reference line.
Meanwhile, we build a 1-D observation for the state information and a 5-D observation for the dynamics model information. Finally, we build a total 32-D observation vector as the representation for each reference line.

For the unicycle model, we set $T_o=10$ and include no dynamics model information. Thus we build a 21-D observation vector.





\section{EXPERIMENTS}

\label{sec:experiments}

In this section, we showcase the results of a series of experiments that validates the proposed method.

\subsection{Experiment Settings}

\subsubsection{Evaluation Method}


We observe that the random walk has a small probability of generating unrealistic data (outliers) and may make the mean of average tracking error perturbed heavily. To handle this problem, we first use Monte-Carlo ~\cite{liu2022learning} to generate a large number of stochastic results. And then, we sort average tracking errors and calculate the median of average tracking errors (see definition in Section~\ref{subsec:trajectory-tracking}) and use it as our metric. 
The metric can be formulated as below:
\begin{equation}
    J_{\text{total}} = \text{median} ( \{ |J_{i} \mid i= 1, 2, ..., N_\text{mc} \} ), \\
\end{equation}
where $N_\text{mc}$ denotes the number of Monte-Carlo simulations for our studies, $i$ denotes the index of simulation, and $J^{(i)}$ is the sample-level metric which will be introduced in the next section. We set $N_\text{mc} = 500$ for our experiments.


\subsubsection{Metrics}


As for the sample (trajectory)-level metric, we use the average $L^2$ distance between the reference line and the tracked trajectory. This metric measures the general tracking quality of one pass of tracking and is invariant to the sequence length of the reference line. The sample-level metric can be indicated as
\begin{equation}
\begin{array}{c}
    \mathbf{z}^{\star} = \{ z_{t}^{\star} \} \sim \mathcal{Z} \\
    J_{i} = \frac{1}{T} \sum_{t=0}^{T-1} \Vert z_{t} - z_{t}^{\star} \Vert_2 \\
\end{array},
\end{equation}
 where $\mathcal{Z}$ denotes the distribution which reference lines are sampled from for each stochastic simulation.

\subsubsection{Baseline and Post-Optimizer}
We choose the pure pursuit~\cite{coulter1992implementation} tracker as our baseline as it is a typical heuristics-based method and has a wide range of applications in the autonomous driving industry.
The pure pursuit method determines the steering angle based on a look-ahead point on the reference path and executes this steering angle to roll out a curve that can reach the look-ahead point. As for longitudinal/speed control, we build a proportional controller~\cite{10.5555/2675399}.

The trajectory optimization is built with iLQR~\cite{li2004iterative} to refine the tracking result further. Specifically, we formulate the optimal control problem as a quadratic programming problem with equality constraints at each iteration.

\subsection{Main Results}

\label{subsec:main-results}


\def\InnerExpGroupDivider{\cline{4-12}}
\def\InterExpGroupDivider{\specialrule{.2em}{.1em}{.1em} }
\def\InterSettingGroupDivider{\hline\hline}

\newcommand{\MarkCmpParam}[1]{\underline{#1}}  

\newcommand{\MarkExpBest}[1]{$\mathbf{#1}$}
\newcommand{\MarkExpTrivialBest}[1]{$\mathit{#1}$}
\newcommand{\MarkSettingBest}[1]{$\mathbf{#1}^{\star}$}
\newcommand{\MarkSettingTrivialBest}[1]{$\mathit{#1}^{\star}$}

\begin{table*}
\vspace{+0.2cm}  
\caption{
Comparison between different trajectory tracking methods.
\newline The best result of each iteration is marked in bold if having a 5\%+ relative error reduction compared to the opponent, otherwise in italics.
\newline The best result of each setting (initial speed/noise level/dynamics model) is marked with a star.
\newline Parameters modified for comparison are underlined.
}
\label{table:compare-trackers}
\begin{center}
\resizebox{1.0\textwidth}{!}{
\begin{tabular}{|c|c|c|c|c|c|c|c|c|c|c|c|}  
\hline
\multicolumn{4}{|c|}{Optimization iteration} & 
\multicolumn{2}{c|}{0} &
\multicolumn{2}{c|}{1} &
\multicolumn{2}{c|}{2} &
\multicolumn{2}{c|}{3} 
\\ 
\hline
\makecell{Initial \\ speed \\ $v_\text{init}$ } & \makecell{Noise \\ level \\ $w_\text{n}$} & \makecell{Dynamics \\ model \\ $\mathbf{T}$} & Method
& Error & \makecell{Relative \\ error \\ change}
& Error & \makecell{Relative \\ error \\ change}
& Error & \makecell{Relative \\ error \\ change}
& Error & \makecell{Relative \\ error \\ change}
\\
\InterSettingGroupDivider


\multirow{3}*{\MarkCmpParam{5.0 [m/s]}} & \multirow{3}*{0.0} & \multirow{3}*{\makecell{Bicycle \\ model}} & pp+ilqr & 0.2555& 0.0000   &   0.1905& 0.0000   &   0.0246& 0.0000   &   \MarkExpTrivialBest{0.0280}& 0.0000
\\
\InnerExpGroupDivider
~ & ~ & ~ & drl+ilqr & 0.0981&-0.6158   &   0.1321&-0.3064   &   \MarkSettingBest{0.0220}&-0.1038   &   0.0291& 0.0392
\\
\InnerExpGroupDivider
~ & ~ & ~ & drl-L+ilqr & \MarkExpBest{0.0605}&-0.7634   &   \MarkExpBest{0.0954}&-0.4993   &   0.0224&-0.0894   &   0.0297& 0.0635
\\
\InterExpGroupDivider

\multirow{3}*{\MarkCmpParam{10.0 [m/s]}} & \multirow{3}*{0.0} & \multirow{3}*{\makecell{Bicycle \\ model}} & pp+ilqr & 0.2467& 0.0000   &   0.2973& 0.0000   &   0.0254& 0.0000   &   \MarkSettingTrivialBest{0.0219}& 0.0000
\\
\InnerExpGroupDivider
~ & ~ & ~ & drl+ilqr & 0.1379&-0.4409   &   0.2874&-0.0332   &   0.0235&-0.0727   &   0.0220& 0.0049
\\
\InnerExpGroupDivider
~ & ~ & ~ & drl-L+ilqr & \MarkExpBest{0.0804}&-0.6742   &   \MarkExpBest{0.1676}&-0.4363   &   \MarkExpBest{0.0220}&-0.1344   &   0.0220& 0.0039
\\
\InterExpGroupDivider

\multirow{3}*{\MarkCmpParam{15.0 [m/s]}} & \multirow{3}*{0.0} & \multirow{3}*{\makecell{Bicycle \\ model}} & pp+ilqr & 0.2927& 0.0000   &   0.5505& 0.0000   &   0.0292& 0.0000   &   0.0189& 0.0000
\\
\InnerExpGroupDivider
~ & ~ & ~ & drl+ilqr & 0.1567&-0.4646   &   0.4770&-0.1335   &   0.0255&-0.1258   &   \MarkSettingTrivialBest{0.0188}&-0.0011
\\
\InnerExpGroupDivider
~ & ~ & ~ & drl-L+ilqr & \MarkExpBest{0.0867}&-0.7037   &   \MarkExpBest{0.2483}&-0.5490   &   \MarkExpBest{0.0213}&-0.2716   &   0.0189&-0.0007
\\
\InterExpGroupDivider

\multirow{3}*{\MarkCmpParam{20.0 [m/s]}} & \multirow{3}*{0.0} & \multirow{3}*{\makecell{Bicycle \\ model}} & pp+ilqr & 0.4239& 0.0000   &   1.0195& 0.0000   &   0.0427& 0.0000   &   0.0181& 0.0000
\\
\InnerExpGroupDivider
~ & ~ & ~ & drl+ilqr & 0.1753&-0.5866   &   0.7526&-0.2618   &   0.0310&-0.2722   &   0.0181&-0.0031
\\
\InnerExpGroupDivider
~ & ~ & ~ & drl-L+ilqr & \MarkExpBest{0.1007}&-0.7625   &   \MarkExpBest{0.4378}&-0.5706   &   \MarkExpBest{0.0219}&-0.4873   &   \MarkSettingTrivialBest{0.0180}&-0.0047
\\
\InterExpGroupDivider

\multirow{3}*{\MarkCmpParam{25.0 [m/s]}} & \multirow{3}*{0.0} & \multirow{3}*{\makecell{Bicycle \\ model}} & pp+ilqr & 0.6468& 0.0000   &   1.7095& 0.0000   &   0.0929& 0.0000   &   0.0217& 0.0000
\\
\InnerExpGroupDivider
~ & ~ & ~ & drl+ilqr & 0.2115&-0.6731   &   1.0289&-0.3981   &   \MarkExpBest{0.0475}&-0.4889   &   0.0217& 0.0004
\\
\InnerExpGroupDivider
~ & ~ & ~ & drl-L+ilqr & \MarkExpBest{0.1457}&-0.7747   &   \MarkExpBest{0.9251}&-0.4589   &   0.0499&-0.4628   &   \MarkSettingTrivialBest{0.0216}&-0.0008
\\
\InterExpGroupDivider

\multirow{3}*{\MarkCmpParam{30.0 [m/s]}} & \multirow{3}*{0.0} & \multirow{3}*{\makecell{Bicycle \\ model}} & pp+ilqr & 0.9755& 0.0000   &   3.3988& 0.0000   &   0.6060& 0.0000   &   0.2761& 0.0000
\\
\InnerExpGroupDivider
~ & ~ & ~ & drl+ilqr & 0.2774&-0.7156   &   \MarkExpBest{2.0135}&-0.4076   &   \MarkExpBest{0.2683}&-0.5573   &   \MarkSettingBest{0.2565}&-0.0709
\\
\InnerExpGroupDivider
~ & ~ & ~ & drl-L+ilqr & \MarkExpBest{0.2688}&-0.7245   &   2.5946&-0.2366   &   0.3582&-0.4090   &   0.2567&-0.0704
\\
\InterSettingGroupDivider


\multirow{3}*{20.0 [m/s]} & \multirow{3}*{ \MarkCmpParam{0.003} } & \multirow{3}*{\makecell{Bicycle \\ model}} & pp+ilqr & 0.4159& 0.0000   &   0.9766& 0.0000   &   0.0541& 0.0000   &   \MarkSettingTrivialBest{0.0234}& 0.0000
\\
\InnerExpGroupDivider
~ & ~ & ~ & drl+ilqr & 0.1761&-0.5767   &   0.7155&-0.2673   &   0.0390&-0.2785   &   0.0235& 0.0027
\\
\InnerExpGroupDivider
~ & ~ & ~ & drl-L+ilqr & \MarkExpBest{0.1009}&-0.7573   &   \MarkExpBest{0.4413}&-0.5481   &   \MarkExpBest{0.0298}&-0.4482   &   0.0235& 0.0051
\\
\InterExpGroupDivider

\multirow{3}*{20.0 [m/s]} & \multirow{3}*{ \MarkCmpParam{0.01} } & \multirow{3}*{\makecell{Bicycle \\ model}} & pp+ilqr & 0.4193& 0.0000   &   1.1238& 0.0000   &   \MarkExpTrivialBest{0.3760}& 0.0000   &   \MarkExpTrivialBest{0.3698}& 0.0000
\\
\InnerExpGroupDivider
~ & ~ & ~ & drl+ilqr & 0.1769&-0.5782   &   0.8995&-0.1996   &   0.3817& 0.0150   &   0.3713& 0.0039
\\
\InnerExpGroupDivider
~ & ~ & ~ & drl-L+ilqr & \MarkSettingBest{0.1025}&-0.7555   &   \MarkExpBest{0.6508}&-0.4209   &   0.3864& 0.0276   &   0.3761& 0.0170
\\
\InterExpGroupDivider

\multirow{3}*{20.0 [m/s]} & \multirow{3}*{ \MarkCmpParam{0.03} } & \multirow{3}*{\makecell{Bicycle \\ model}} & pp+ilqr & 0.4231& 0.0000   &   3.2279& 0.0000   &   2.8892& 0.0000   &   \MarkExpTrivialBest{2.9015}& 0.0000
\\
\InnerExpGroupDivider
~ & ~ & ~ & drl+ilqr & 0.1776&-0.5802   &   \MarkExpBest{3.0604}&-0.0519   &   \MarkExpTrivialBest{2.8520}&-0.0129   &   2.9029& 0.0005
\\
\InnerExpGroupDivider
~ & ~ & ~ & drl-L+ilqr & \MarkSettingBest{0.1079}&-0.7451   &   3.0787&-0.0462   &   2.8764&-0.0044   &   2.9097& 0.0028
\\
\InterSettingGroupDivider


\multirow{3}*{\MarkCmpParam{2.0 [m/s]}} & \multirow{3}*{ 0.0 } & \multirow{3}*{\makecell{\MarkCmpParam{Unicycle} \\ \MarkCmpParam{model}}} & pp+ilqr & 0.1940& 0.0000   &   0.0448& 0.0000   &   0.0135& 0.0000   &   0.0133& 0.0000
\\
\InnerExpGroupDivider
~ & ~ & ~ & drl+ilqr & 0.1008&-0.4805   &   0.0202&-0.5501   &   \MarkSettingTrivialBest{0.0132}&-0.0218   &   \MarkSettingTrivialBest{0.0132}&-0.0107
\\
\InnerExpGroupDivider
~ & ~ & ~ & drl+ilqr & \MarkExpBest{0.0834}&-0.5702   &   \MarkExpBest{0.0171}&-0.6175   &   \MarkSettingTrivialBest{0.0132}&-0.0218   &   \MarkSettingTrivialBest{0.0132}&-0.0107
\\
\hline

\end{tabular}
}
\end{center}
\end{table*}

In this section, we validate the proposed method from accuracy, robustness, and versatility. For accuracy, we prove that the proposed DRL-based trajectory tracking method has an obvious improvement in accuracy compared to the heuristics-based method. For robustness, we prove the DRL-based method is robust to perform the trajectory tracking task under noise. And for versatility, we validate the ability of our method to integrate with multiple dynamics models.

\subsubsection{Accuracy Demonstration within a Wide Initial Speed Range}


To demonstrate the priority on the accuracy, we conduct comparison experiments as shown Table~\ref{table:compare-trackers}. These results demonstrate that the DRL-based trajectory tracker can make a consistent and significant improvement in average tracking error across different optimization steps.

We also observe that the accuracy of our DRL-based method becomes better when the speed gets higher. That is, under a low initial speed such as 5 [m/s], although the direct output at optimization step 0 of the DRL-based tracker has an advantage over the one of the pure-pursuit tracker, this advantage diminished rapidly when the post-optimization goes on (i.e. the iLQR optimization begins to iterate). In contrast, when the initial speed goes higher such as 20 or 30 [m/s], the DRL-based tracker can keep a significant gain at every optimization step (over 5\% relative reduction in tracking error) compared with the heuristics-based tracker. The reason is that the heuristics rely on the assumption that requires relatively steady states. This can easily hold under low speed, but not under high speed.

This observation partially reveals the dilemma of the heuristics-based method. Hand-crafted heuristics are often based on assumptions on motion patterns and may fail in real operation scenarios whose variation can never be fully covered during the design phase. When facing a new application, it is often difficult to identify its suitable application domain, which often requires complex analysis tools such as the reachability analysis~\cite{bansal2017hamilton}. This disadvantage further limits the development and application.

Figure~\ref{figure:comparison-results-different-init_v} presents how the trajectory tracker and the trajectory optimizer cooperate by depicting the variation of tracking errors at different optimization steps. One general observation is that for all tracking methods (including our DRL-based method), the errors can be reduced further via the post-optimization process, except for very high speeds (e.g. 30 [m/s] / 108 [kph]) which is challenging for iLQR optimizer as the dynamics become highly non-linear under this context. Thus the trajectory tracker can serve as a module producing an initial trajectory located as close as possible to the optimum. When this initial trajectory is fed into a trajectory optimizer, it is easier for the refinement to get a more accurate trajectory.


\def\InitVSubfigureWidth{0.22\textwidth}

\begin{figure}
\vspace{+0.2cm}  
\centering
 \begin{subfigure}[b]{\InitVSubfigureWidth}
     \centering
     \includegraphics[width=\textwidth]{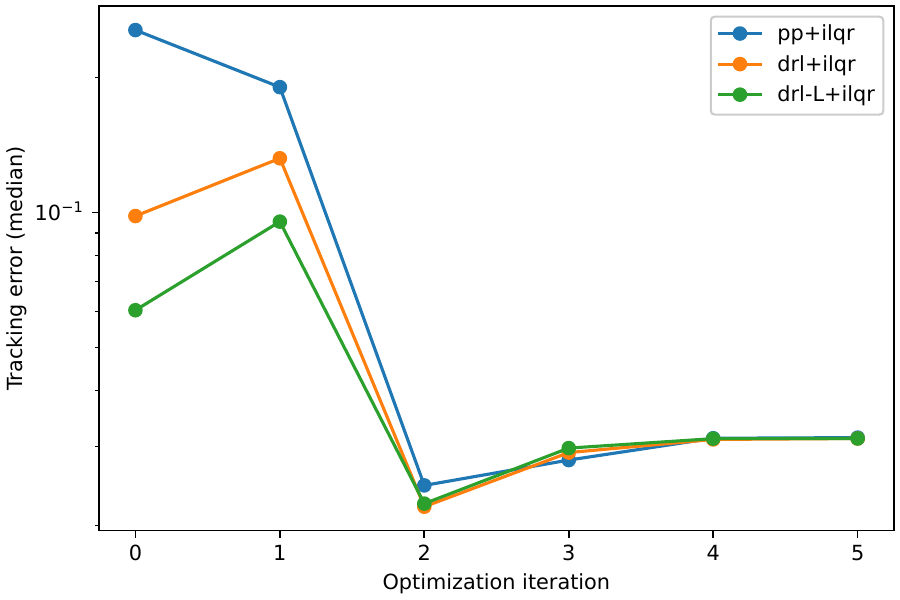}
     \caption{$v_\text{init} = 5$}
 \end{subfigure}
 \hfill
 \begin{subfigure}[b]{\InitVSubfigureWidth}
     \centering
     \includegraphics[width=\textwidth]{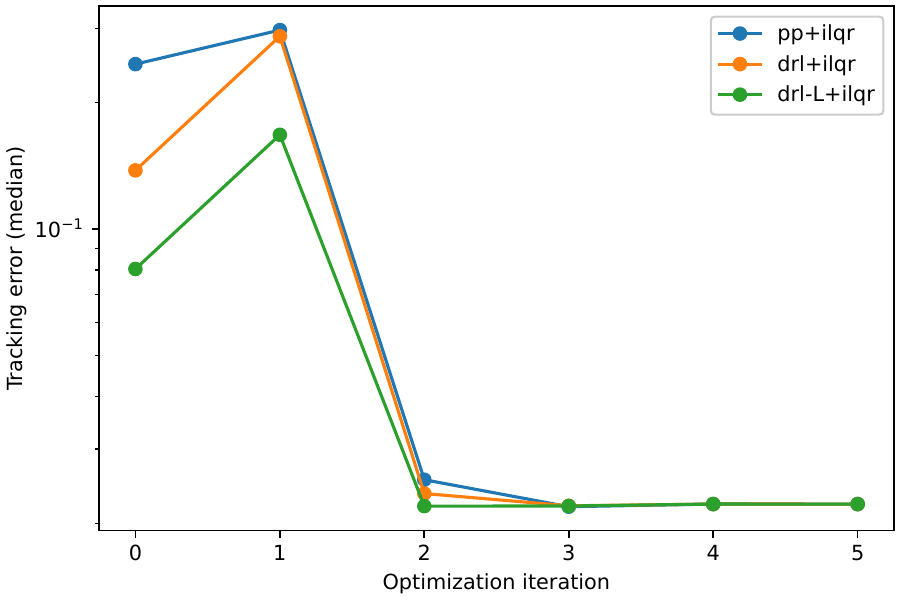}
     \caption{$v_\text{init} = 10$}
 \end{subfigure}
 \hfill
 \begin{subfigure}[b]{\InitVSubfigureWidth}
     \centering
     \includegraphics[width=\textwidth]{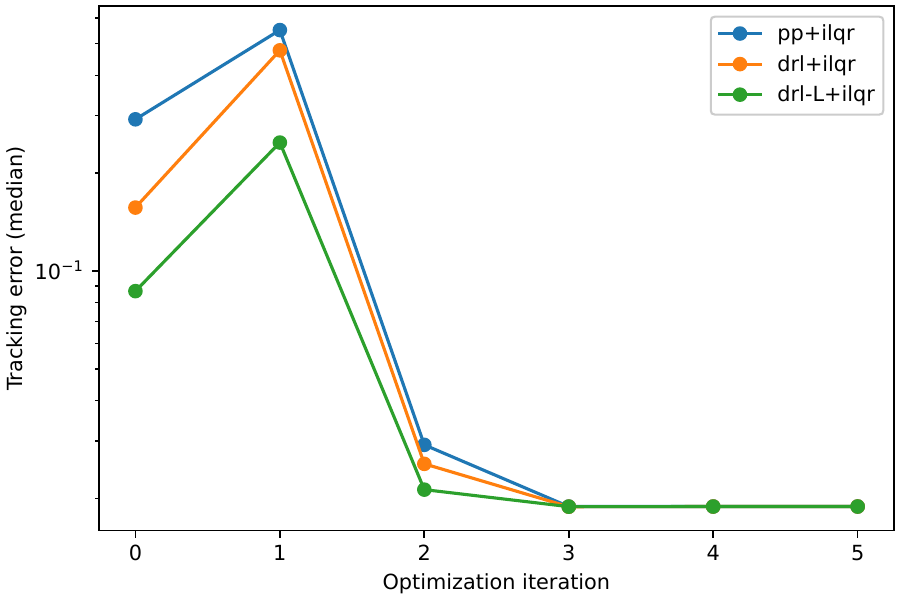}
     \caption{$v_\text{init} = 15$}
 \end{subfigure}
 \hfill
 \begin{subfigure}[b]{\InitVSubfigureWidth}
     \centering
     \includegraphics[width=\textwidth]{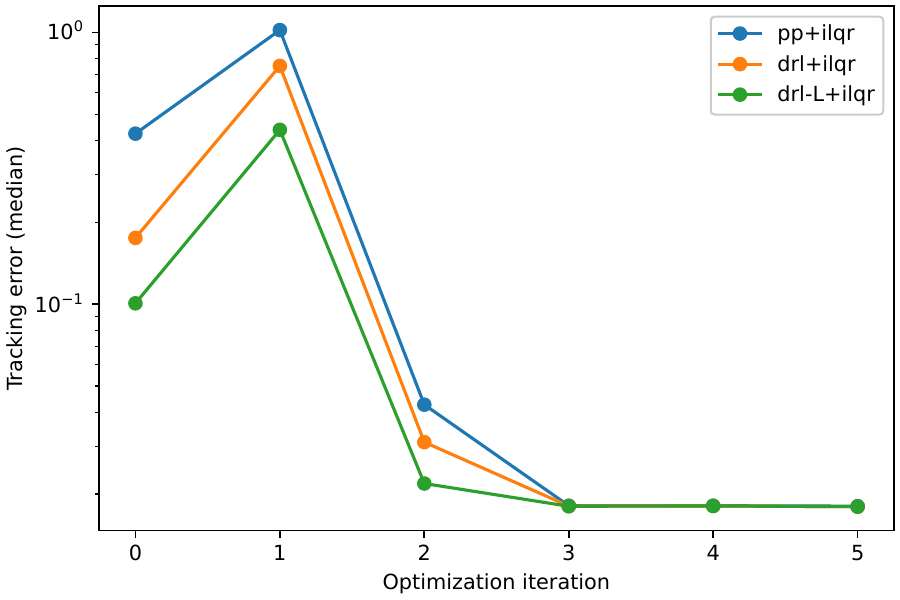}
     \caption{$v_\text{init} = 20$}
 \end{subfigure}
 \hfill
 \begin{subfigure}[b]{\InitVSubfigureWidth}
     \centering
     \includegraphics[width=\textwidth]{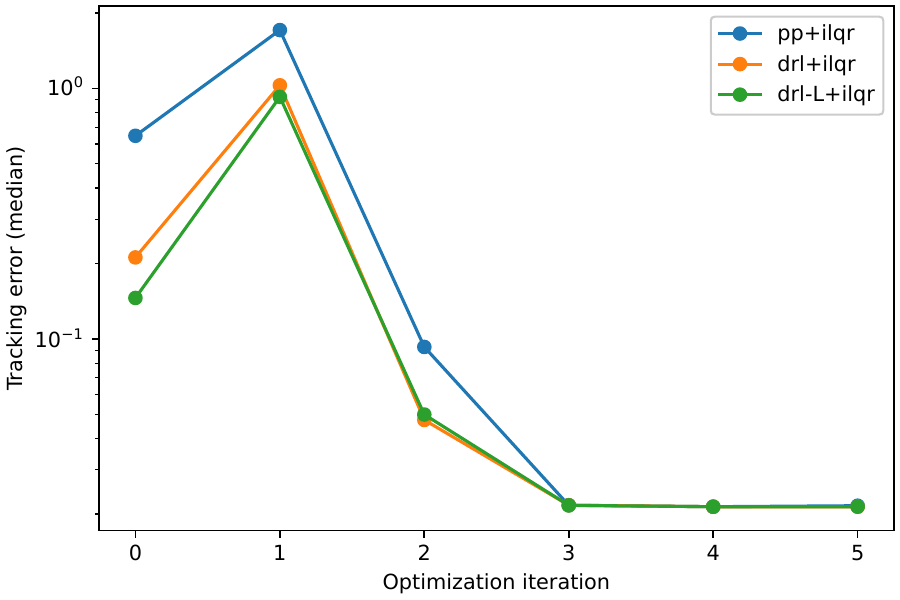}
     \caption{$v_\text{init} = 25$}
     \label{subfugure:init-v-25}
 \end{subfigure}
 \hfill
 \begin{subfigure}[b]{\InitVSubfigureWidth}
     \centering
     \includegraphics[width=\textwidth]{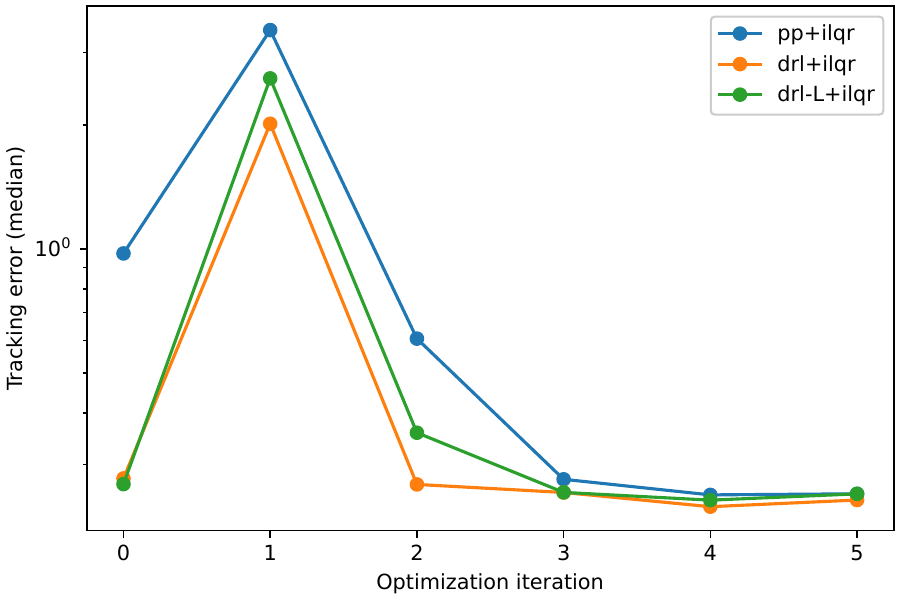}
     \caption{$v_\text{init} = 30$}
 \end{subfigure}

\caption{Trajectory tracking and trajectory optimization results under different initial speeds. Better viewed in color with zoom-in.}
\label{figure:comparison-results-different-init_v}
\end{figure}

\subsubsection{Robustness Demonstration under Noise}

We prove the robustness of our DRL-based method by testing whether it can handle noise that is ubiquitous in real-world applications. To investigate this factor, we consider injecting random a noise signal which has a magnitude that is proportional to the step length within a single time interval. This process can be written as
\begin{equation}
\begin{array}{c}
    \mathbf{z}^{n} \sim \mathcal{N}(\mathbf{0}, \bm{\sigma}), \bm{\sigma} = \sigma \cdot \mathbf{I}, \sigma = v_\text{init} \cdot \delta t \cdot w_\text{n}\\
    \mathbf{z}^{\star} \mapsfrom \{\mathbf{z}_{i}^{\star} + \mathbf{z}_{i}^{n} \mid i = 0, 1, ..., T-1\},  \\
\end{array}
\end{equation}
where $\mathbf{I}$ denotes the identity matrix, $w_\text{n}$ denotes the noise level i.e. amplitude, and the addition operator here is the element-wise addition on coordinates.

Results in Table~\ref{table:compare-trackers} show that when the amplitude of noise raises, post-optimization may encounter failure due to its model-based nature, and the initial tracking results will become the best results. In this case, our DRL-based method yields better results consistently. This observation reveals the potential of the DRL-based method to be applied in real-world applications where noise can often be rooted in sensor hardware, imperfect high-level planners, actuators with limited precision, and so on.

\subsubsection{Versatility Demonstration with Different Dynamics Models}

To show the versatility of the model-free DRL-based method, we conduct additional comparison under the unicycle model~\cite{batkovic2018computationally} which simulates the scenarios with pedestrians. As shown in Table~\ref{table:compare-trackers}, our method demonstrates a consistent improvement in tracking errors.

\subsection{Visualization}

To explain the effectiveness of our method, we present two cases with their learned value functions as heatmaps visualized in Figure~\ref{figure:value-function}. The learned "optimal action-value function" $V^{\ast}(s)$ is the prediction of our value network taking the current state from the heatmap position and current action, while the learned "optimal state-value function" $Q^{\ast}(s, a)$ is the prediction of the same network, which takes action re-predicted by our policy network at each heatmap position with re-encoded observation.

From the heatmap, we observe a region of high value centered at the current reference waypoint, which can be explained as guidance for the policy network to drive the agent toward the current reference waypoint. Concretely, DRL learning methods such as the member from the "actor-critic" family~\cite{fujimoto2018addressing,haarnoja2018soft} use the gradient information from the value network to derive the gradient of reward w.r.t. the action $\frac{\partial V(s, a)}{\partial a}$. Then they propagate it to the policy network and finally get the gradient information to optimize the parameters in the policy network $\theta_\Pi$. With the chain rule, we have
\begin{equation}
    \frac{\partial r}{\partial \theta_{\Pi}} = \frac{\partial r}{\partial a} \frac{\partial a}{\partial \theta_{\Pi}} = \frac{\partial V(s, a)}{\partial a} \frac{\partial \Pi(s)}{\partial \theta_{\Pi}}.
\end{equation}

Additionally, a consistent rise in the value (warmer color) of the optimal state-value function than the optimal action-value function can be observed by comparing two columns in each case. This corresponds to the following relation derived from the definition
\begin{equation}
    V^{\ast}(s) \triangleq \max\limits_{a} Q^{\ast}(s, a) \geq Q^{\ast}(s, a).
\end{equation}

\def\ValueFunctionSubfigureWidth{0.23\textwidth}

\begin{figure}
\vspace{+0.2cm}  
\centering
 \begin{subfigure}[b]{\ValueFunctionSubfigureWidth}
     \centering
     \includegraphics[width=\textwidth]{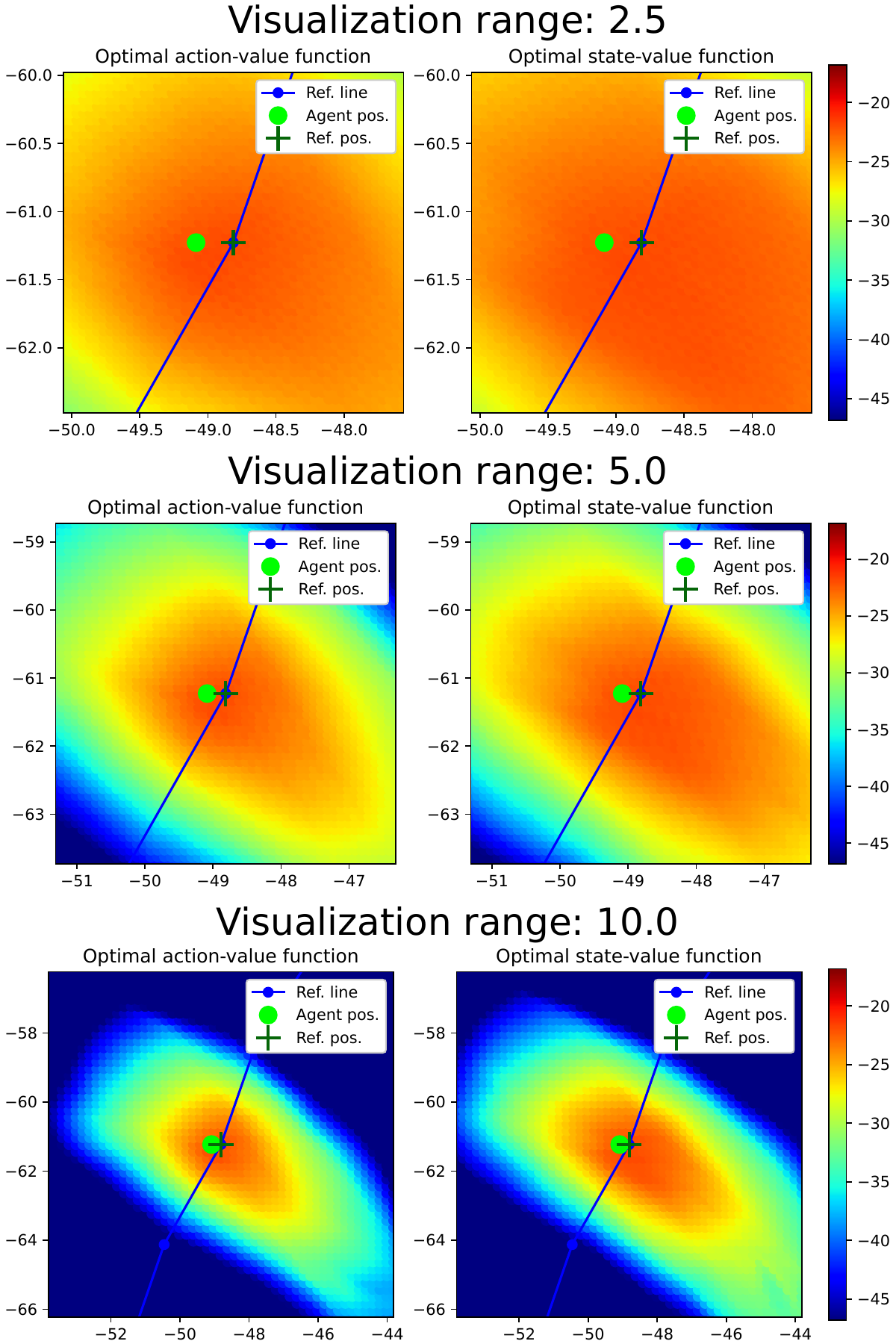}
 \end{subfigure}
 \hfill
 \begin{subfigure}[b]{\ValueFunctionSubfigureWidth}
     \centering
     \includegraphics[width=\textwidth]{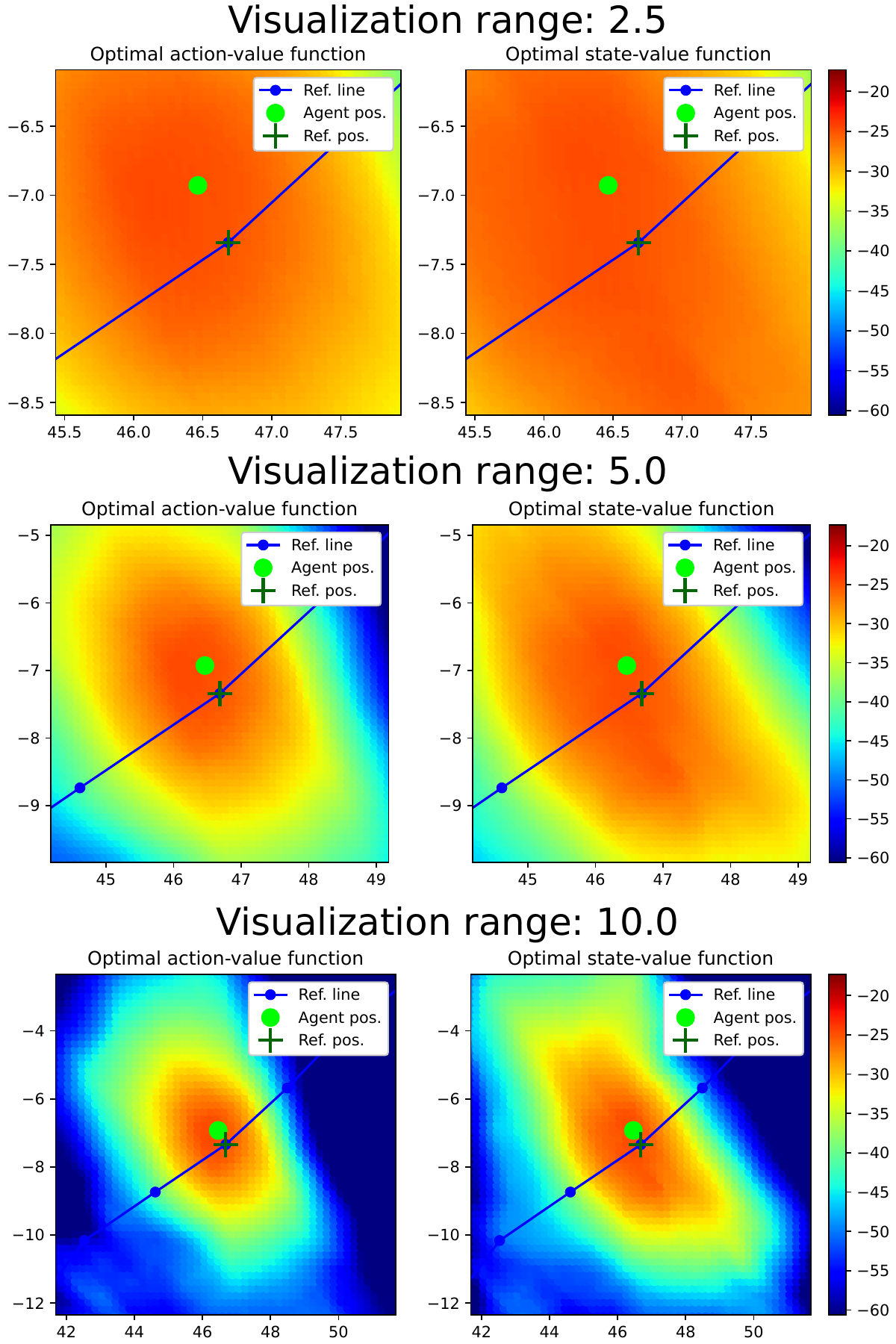}
 \end{subfigure}
\caption{Optimal action-value function and state-value function. Different rows represent different scales. A warmer color denotes a region with a higher estimated accumulated reward in the future, while a colder one represents the opposite. Better viewed in color with zoom-in.}
\label{figure:value-function}
\end{figure}










\section{CONCLUSIONS}

In this paper, we proposed a DRL-based trajectory tracking method. Our method took advantage of both the representation learning ability of NN and the exploration nature of RL to provide stronger robustness. Besides, we made the least assumption on the model and improved accuracy and versatility. From experiments, we demonstrated the priority of our DRL-based method compared to other methods.

\addtolength{\textheight}{-0cm}   










\bibliographystyle{IEEEtran}
\bibliography{./refs}

\end{document}